\documentclass[sigconf]{acmart}
\usepackage[T1]{fontenc}

\AtBeginDocument{%
  \providecommand\BibTeX{{%
    \normalfont B\kern-0.5em{\scshape i\kern-0.25em b}\kern-0.8em\TeX}}}

\setcopyright{none}
\copyrightyear{}
\acmYear{}
\acmDOI{}

\acmConference[DSHealth '22]{DSHealth '22: 2022 KDD Workshop on Applied Data Science for Healthcare}{August 2022}{Washington DC, USA}
\acmBooktitle{}
\acmPrice{}
\acmISBN{}

\usepackage{amsmath}
\usepackage{marginnote}
\usepackage{multicol}
\usepackage{enumitem}
\usepackage{natbib}
\usepackage{xcolor}
\usepackage{booktabs}
\usepackage{adjustbox}
\usepackage{subcaption}
\usepackage{array}
\usepackage{tikz}
\usetikzlibrary{shapes.geometric, arrows, shapes.multipart}

\tikzstyle{startstop} = [rectangle, rounded corners, minimum width=3cm, minimum height=1cm,text centered, draw=blue, fill=blue!30] 
\tikzstyle{interim} = [rectangle, rounded corners, minimum width=3cm, minimum height=1cm,text centered] 
\tikzstyle{io} = [trapezium, trapezium left angle=70, trapezium right angle=110, minimum width=3cm, minimum height=1cm, text centered, draw=black, fill=blue!30]
\tikzstyle{process} = [rectangle, minimum width=3cm, minimum height=1cm, text centered, text width=3cm, draw=black, fill=orange!30]
\tikzstyle{decision} = [diamond, aspect=5, minimum width=3cm, minimum height=1cm, text centered, draw=black] 
\tikzstyle{arrow} = [thick,->,>=stealth]

\begin{document}

\newcommand{\todotip}[2][NA]{{\color{red} $\rhd${[Assignee #1]: #2}}}
\newcommand{\explannote}[2][NA]{{\reversemarginpar \color{blue} $\lhd${[#1 says]: #2}}}

\newcolumntype{A}{>{\raggedright}p{0.75\linewidth}}
\newcolumntype{B}{>{\raggedright}p{0.25\linewidth}}

\title{Distillation to Enhance the Portability of Risk Models Across Institutions with Large Patient Claims Databases}

\newcommand\Mark[1]{\textsuperscript{#1}}
\author{
        Steve Nyemba\Mark{1},
        Chao Yan\Mark{1},
        Ziqi Zhang\Mark{2},
        Amol Rajmane\Mark{3},
        Pablo Meyer\Mark{4},\\
        Prithwish Chakraborty\Mark{4},
        Bradley Malin\Mark{1,2}
      }
\affiliation{
  \institution{
    \Mark{1} Vanderbilt University Medical Center, Nashville, TN, USA\\
    \Mark{2} Vanderbilt University, Nashville, TN, USA\\
    \Mark{3} IBM Watson Health, MA, USA\\
    \Mark{4} Center for Computational Health, IBM Research, NY, USA\\
    }
  \city{ }
  \country{ }
  }

\renewcommand{\shortauthors}{Nyemba, et al.}

\begin{abstract}

Artificial intelligence, and particularly machine learning (ML), is increasingly developed and deployed to support healthcare in a variety of settings. However, 
clinical decision support (CDS) technologies based on ML need to be portable if they are to be adopted on a broad scale.
In this respect, models developed at one institution should be reusable at another.
Yet there are numerous examples of portability failure, 
particularly due to naive application of ML models. Portability failure can lead to  
suboptimal care and medical errors, which ultimately could  prevent the adoption of ML-based CDS in practice.
One specific healthcare challenge that could benefit from enhanced portability is
the prediction of 30-day readmission risk. Research to date has shown that deep learning models
can be effective at modeling such risk.
In this work, we investigate the practicality of model portability through a cross-site evaluation of readmission prediction models. To do so, we apply a recurrent neural network, augmented with self-attention and blended with expert features,
to build readmission prediction models for two independent large scale claims datasets. 
We further present a novel transfer learning technique that adapts the well-known method of born-again network (BAN) training.
Our experiments show that direct application of ML models trained at one institution and tested at another institution perform worse than models trained and tested at the same institution.
We further show that the transfer learning approach based on the BAN 
produces models that are better than those trained on just a single institution's data. Notably, this improvement is consistent across both sites and occurs after a single retraining, which illustrates the potential for a cheap and general model transfer mechanism of readmission risk prediction.

\end{abstract}

\ccsdesc[500]{Computing methodologies~Artificial intelligence}
\ccsdesc[500]{Applied computing~Health informatics}
\ccsdesc[300]{Computing methodologies~Learning paradigms}

\keywords{ai for health, model portability, transfer learning}


\maketitle

\section{Introduction}

The quantity of patient data continues to grow at an unprecedented pace. This is supported, in part, by the continued adoption of electronic health record (EHR) systems and use of greater detail in insurance claims. The large scale nature of such resources provide an opportunity to develop artificial intelligence, and machine learning (ML) in particular, to solve problems in a variety of settings. 

From a patient health perspective, ML applications have shown promise for the analysis of
radiology images~\cite{sorantin2021augmented}, modeling co-morbidity of chronic disorders~\cite{kwon2022progression,john2021survey,dey2021impact}, and forecasting of risk in developing particular diseases~\cite{zhang2022}.  
%
From a healthcare administration perspective, there is an expectation that ML can help detect and prevent costly healthcare events, such as the prediction of all-cause 30-day readmission after discharge from a hospital stay. This problem holds real world importance with both societal (preventing
readmissions
reduce the quality of life for patients~\cite{horowitz1}) and economic implications~\cite{fonarow2017hospital}. National payment agencies, such as the U.S. Centers for Medicare and Medicaid Services (CMS), regularly monitor the rate of such an event across healthcare facilities and penalize 
for unplanned readmissions, 
There has been a non-trivial amount of research into 
summary indicators of readmission~\cite{horowitz1}. For instance, Chakraborty et al.~\cite{chakraborty2021blending} showed that effective ML models for forecasting 30-day readmission risk can be learned by jointly representing the complex, sparse, and potentially noisy sequential patient records along with the summary indicators.

Demonstration of a 
model on a dataset from  a single healthcare organization 
is a critical first step in model development.  However, it is important that models developed using data from one site be tested at, and adapted to, other sites.
 Yet there are various challenges that 
can lead to an ineffective transfer
of ML models.
First, there are often inconsistencies in how data is organized and recorded across healthcare organizations. Second, ML models may be overfit to disease patterns observed at a specific site. While this can arise from insufficient model evaluation, patient populations at disparate healthcare organizations may be significantly different.
These problems were clearly illustrated by a failure in portability of Epic's sepsis prediction tool~\cite{Habib2021}. Community efforts, such as the OHDSI network~\cite{hripcsak2015observational}, have worked to solve the first problem by promoting common data model formats across institutes; however, the second aspect is often understudied and is thus the focus of this paper.

In this work, we investigate the practicality of model portability through a cross-site evaluation of readmission prediction models. To do so, we apply a recurrent neural network, augmented with self-attention and blended with expert features,
to build readmission prediction models for two independent large scale claims datasets. Interestingly, it is often desirable from a data-privacy and data-usage restrictions to allow transfer of built models between datasets without actually sharing the underlying training data/gradients. Thus, our focus is on such `black-box' transfer of models (in contrast to federated learning) where we transfer only the final trained model from one dataset and apply it directly on a held-out test data from the second dataset. 
Our experiments show that such `remote' models perform worse than `local' models that are trained on the training data from the second dataset.
Thus, 
we introduce a novel transfer learning technique by adapting the Born-again network (BAN) training paradigm~\cite{furlanello2018born}. BAN is a cheap training routine where the network is trained sequentially over generations, such that soft-decisions on training examples from the model of the previous generation are utilized to achieve more effective models. 
This approach has been found to be effective across a number of domains and applications~\cite{kodialam2021deep}. However, to the best of our knowledge, this 
is the first to adapt this strategy to transfer models across health institutes. There 
specific contributions of this paper are:

\begin{itemize}
  \item We introduce a novel, computationally inexpensive, and effective model transfer strategy based on the BAN training paradigm. It can consider a remote ML model as a black box (in contrast to federated learning/data sharing scenarios where regulated healthcare datasets may need to be explicitly/implicitly combined) while still leading to an effective transfer across institutions.  
  \item We report on a 30-day readmission prediction across two data marts with varied data coverage patterns.  We show that the new strategy can lead to more effective models - achieving over 10\% improvement in the area under the precision-recall curve regardless of the transfer direction (i.e., site A to site B and vice versa) over directly applying remote models and about 2\% improvement over local models.
  \item We show the robustness of this process by analyzing the performance lift of the BAN-transferred remote model compared to the corresponding local model for various bootstrap fold and find consistent improvement over the local models.
\end{itemize}

The remainder of the paper is organized as follows. Next, we describe the experimental setup, including a brief description of the data sets used in this study and the prediction problem. Then, we provide a brief overview of the BAN-driven transfer learning strategy. Finally, we report our results and provide a discussion on future opportunities.

\begin{figure}[htpb]
  \centering
  \resizebox{0.9\linewidth}{!}{
%
%
\begin{tikzpicture}[node distance=2cm, baseline={(current bounding box.east)}]

\node (start) [startstop] {All available claims};
\node (dec1) [decision, below of=start, yshift=0.5cm, ] {Long-term stays?};
\node (dec2) [decision,  text width=4cm, below of=dec1, yshift=-0.5cm] {In-hospital death / transfer to acute / left AMA?};
\node (dec3) [decision,  text width=3cm, below of=dec2, yshift=-0.5cm] {Received treatment for cancer?};
\node (dec4) [decision,  text width=4cm, below of=dec3, yshift=-0.5cm] {Visited Rehab / primary DX for rehab?};
\node (cohort1) [interim,  draw=black, below of=dec4] {Starting Cohort};
\node (dec5) [decision, aspect=3, text width=3cm, below of=cohort1, yshift=-0.5cm] {Additional Claim within 2 days?};
\node (cohort2) [startstop, below of=dec5] {Final Cohort};

\node (dec6) [process, right of=cohort1, xshift=5cm] {30 day readmission flag};
\node (dec7) [decision, aspect=3, text width=3cm, below of=dec6, yshift=-0.5cm] {Is Acute and unplanned proc?};
  \node (cohort3) [process, draw=red, below of=dec7] {Unplanned readmission (Flag)};

  \draw [arrow] (start) -- (dec1);
  \draw [arrow] (dec1) -- node[anchor=east, 
  ] {N}(dec2);
  \draw [arrow] (dec2) -- node[anchor=east, 
  ] {N}(dec3);
  \draw [arrow] (dec3) -- node[anchor=east, 
  ] {N}(dec4);
  \draw [arrow] (dec4) -- node[anchor=east, 
  ] {N}(cohort1);

  \draw [arrow] (cohort1) -- (dec5);
  \draw [arrow] (dec5) -- node[anchor=east] {N}(cohort2);

  \draw [arrow] (cohort1) -- (dec6);
  \draw [arrow] (dec6) -- (dec7);
  \draw [arrow] (dec7) -- node[anchor=east] {Y}(cohort3);

\end{tikzpicture}
%
  }
  \caption{The cohort construction process to study all-cause 30-day readmission.}
  \label{fig:cohort}
\end{figure}
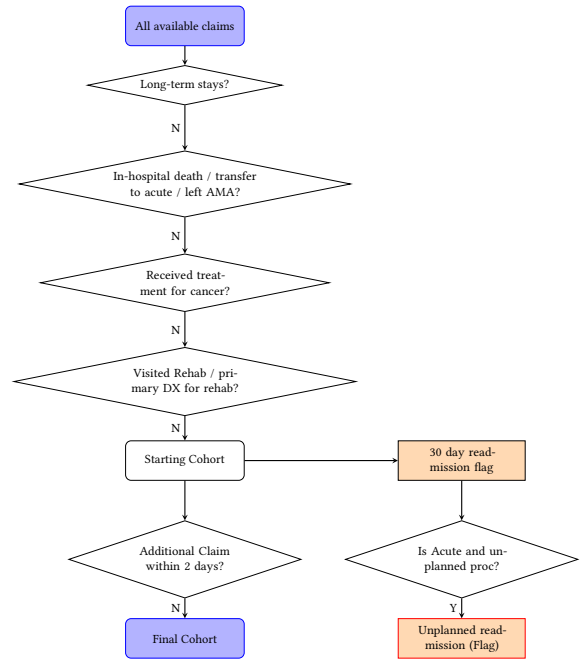

\begin{figure*}[htpb]
  \centering
  \includegraphics[width=0.92\linewidth]{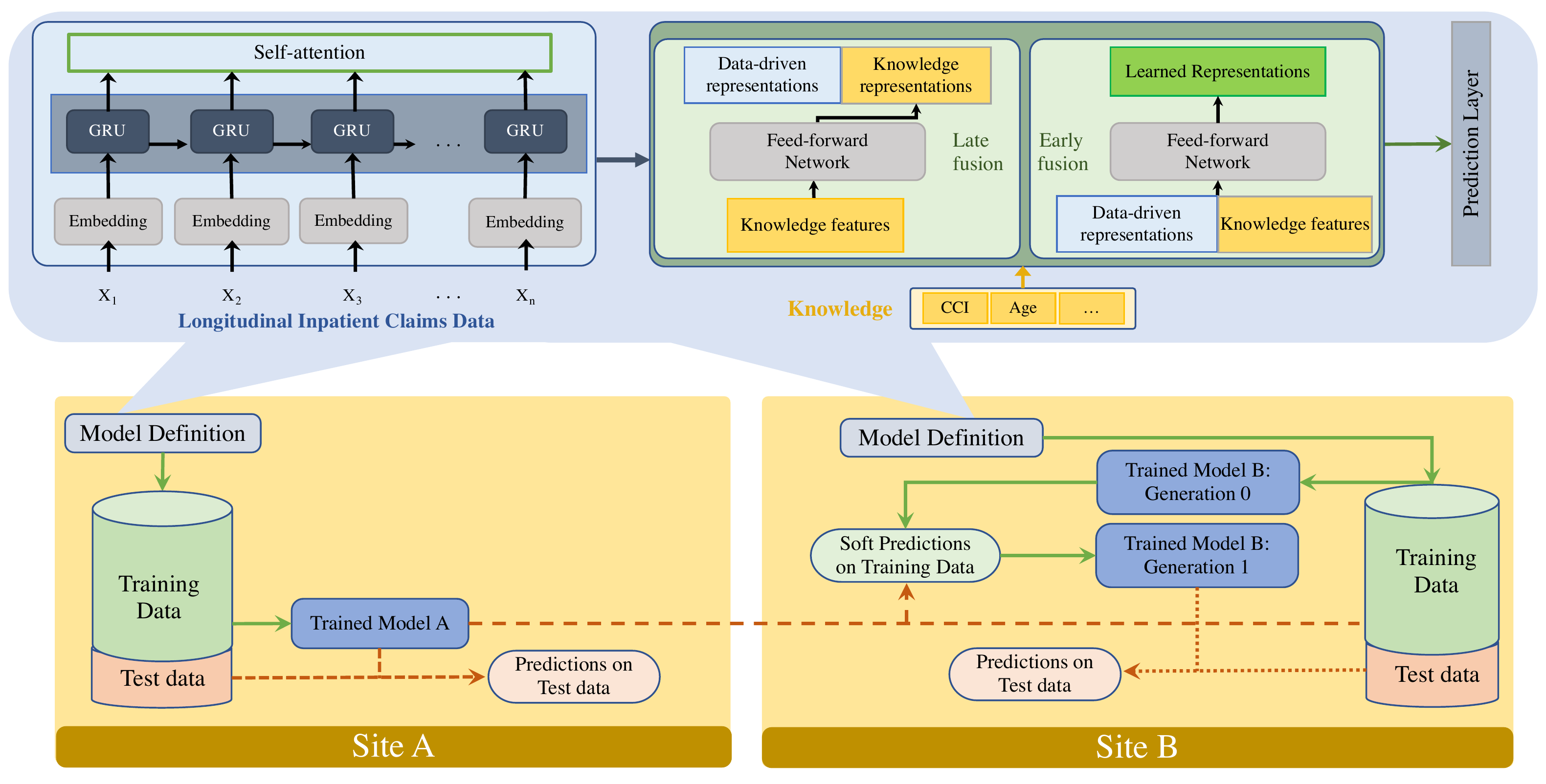}
  \caption{BAN Transfer: Depiction of transfer learning using Born-again-network training. Models share the same definition. Model trained at Site A is used to generate soft-predictions on training data for site B. A model trained on site B is then fine-tuned on these soft-predictions to generate the final transferred model for site B.
  This process implicitly transfers the knowledge from site A in the form of soft decision boundary. 
  Green lines denotes training whereas dotted red-line indicate application of model}%
  \label{fig:training}
\end{figure*}


\section{Experimental Setup}
We conducted the analysis on two separate datasets, specifically the IBM Marketscan Commercial database (IBM Marketscan~\cite{butler2021ibm}) and Vanderbilt University Medical Center (VUMC) data mart, over the same calendar period (2017-2019). IBM Marketscan is intended to be nationally representative, albeit with different geographical sampling, and covering more than 220 million patient lives over the US. By contrast, the VUMC data mart covers substantially fewer patient lives - but it provides more detail on a specific geographic area.  Both datasets cover various data sources such as diagnoses, drugs, and laboratory reports from administrative claims. The Marketscan dataset covers more than 220 million patient lives across the entire United States of America. 
Diagnoses, procedures, and drugs are encoded using the International Classification of Diseases (ICD), Current Procedural Terminology (CPT), and National Drug Code (NDC) terminologies, respectively~\cite{butler2021ibm}. 
The second dataset is from the VUMC, where the data mainly comes from Electronic Data Interchange (EDI) x12 387 reports.
The VUMC data is composed of patients from all fifty states,
with a heavy skew towards the middle Tennessee region. The EDI data is transactional by nature and is, at times, redundant or contradictory as a result of interactions between providers and various payers. Thus, we relied upon data from Vanderbilt's EHR, structured as Observational Medical Outcomes Partnership (OMOP), to consolidate
claim records into a single event per patient. 
The EDI reports enable the VUMC data be semantically consistent with the IBM Marketscan data. 

We constructed all-cause 30-day readmission prediction models following the same cohort definition and considerations as described by Chakraborty \textit{et al.}~\cite{chakraborty2021blending} 
. More specifically, Figure~\ref{fig:cohort} depicts the constructed cohort where we applied several exclusion criteria such as cancer patients, patients who left the hospital against medical advice, and patients who were in rehab. These criteria were aimed at analyzing a standard set of patients for whom we may expect a similar readmission pattern.
%
%
%
%
%
Using this definition, starting from about $220$ million patient lives, we were able to extract $1,112,958$ patients with more than $10$ million records from IBM Marketscan dataset. Similarly, from a starting point of about $5$ million patient records, we were able to extract $1,472,891$ records distributed across $53,810$ patients from the VUMC dataset.
For each of the selected patients, we next defined their `index events' (see ~\cite{chakraborty2021blending}) as the discharge date of a single hospital care or the last date of a set of contiguous hospital care. We also can label each such `index events' with the ground truth label for 30-day readmission. The final dataset is imbalanced with respect to target variables with less than $10\%$ positive labels for each `index event'. 
We extracted a broad set of features for each patients including demographic, medical, and hand-crafter features such as CCI index~\cite{chakraborty2021blending}. 
Finally, we split the data using a $70:10:5:15$ split for training, validation, calibration, and test respectively using a stratified sampling approach. For each dataset, we trained local models with the same architectures as in ~\cite{chakraborty2021blending} using lightsaber, an open source model training framework~\cite{up:suryanarayanan2021disease} to ensure reproduciblity and auditability. The best trained models were selected by running hyperparameter search over the validation set and further calibrated using the calibration set. The test set was used as the holdout and, in the next section, all performance scores are reported based on this set for each dataset.


\section{Methods}
  \label{sec:methods}
Here, we provide an overview of the architecture for model transfer, the process for which is illustrated in Figure~\ref{fig:training}.  Let us represent the data for two datasets as $\mathcal{D}_1: \lbrace X_1, Y_1 \rbrace$ and $\mathcal{D}_2: \lbrace X_2, Y_2 \rbrace$, 
where $X_{i} = \{x_{1i}, x_{2i}, \cdots x_{N_{i}i}\}$ denote the set of features for $N_{i}$ index-events for the dataset $\mathcal{D_{i}}$ and
$Y_{i}=\{y_{1i}, y_{2i}, \cdots y_{N_{i}i}\}$ is a set of boolean values represent the corresponding ground truth for 30-day readmission. 
Furthermore, let us denote the local models trained on their respective dataset as $\mathcal{M}_1$ and $\mathcal{M}_2$. These are referred to as the local models when used in context for the respective datsets. In contrast, we refer to these as the remote models when using it for the alternate dataset e.g. $\mathcal{M}_1$ used to predict readmission on $\mathcal{D}_2$.

Without loss of generality, we now describe the transfer process by considering models are trained on $\mathcal{D}_1$ and applied on $\mathcal{D}_2$. 
With respect to $\mathcal{D}_2$, its local model $\mathcal{M}_2$ is trained by minimizing the cross-entropy loss on the corresponding training instances as:
\begin{equation}
  L(\mathcal{D}_2) = L(\hat{y}_2, y_2)
\end{equation}
where $L$ is the cross-entropy loss, $y_2$ is the target value in the training instances from dataset $\mathcal{D}_2$, and $\hat{y}_2$ is the soft-decision score ($\in [0, 1]$) output by $\mathcal{M}_2$ for training instances $\{x_{j2}; j \in \text{train-set}(\mathcal{D}_2\}$.

In the proposed BAN-transfer procedure, this local model is further fine-tuned using outputs $\hat{y}_{1 \rightarrow 2}$ predicted on data mart $\mathcal{D}_2$ using remote model $\mathcal{M}_1$ (trained on $\mathcal{D}_1$).  
The final model, hereafter referred to as the BAN-transferred model, 
is then trained by fine-tuning $\mathcal{M}_2$ and minimizing the cross-entropy loss as below:
\begin{equation}
  L(\mathcal{D}_2, \mathcal{M}_1) = L(\hat{y}_2, \hat{y}_{1 \rightarrow 2})
\end{equation}

\section{Results}
  \label{sec:results}
  \begin{table*}
  \centering
  \caption{Performance evaluation of the locally learned and transfer learned models per site.}
  \label{tab:perf}
  \begin{tabular}{lllll}
  \toprule
                        & \multicolumn{2}{c}{IBM}   & \multicolumn{2}{c}{VUMC} \\
                        \cmidrule(lr){2-3}          \cmidrule(lr){4-5} 
  Model                 & AUROC            & AUPRC              & AUROC             & AUPRC               \\
  \cmidrule(lr){1-1} \cmidrule(lr){2-5}
  IBM                   & 0.585 (baseline) & 0.290  (baseline)   & 0.600 (-1.15$\%$) & 0.259  (-13.67$\%$) \\
  VUMC                  & 0.610 (4.3$\%$) & 0.260  (-10.34$\%$) & 0.607 (baseline)  & 0.300   (baseline)   \\
  \cmidrule(lr){1-1} \cmidrule(lr){2-5}
  BAN transferred VUMC  & 0.611 (4.4$\%$) & 0.295 (1.72$\%$)   & -                 & -                   \\
  BAN transferred IBM    &   -              &  -                 & 0.614 (1.15$\%$)  & 0.308    (2.6 $\%$)    \\
  \bottomrule
\end{tabular}

\end{table*}

\begin{figure}[h]
  \centering
  \begin{subfigure}{.47\textwidth}
    \centering
    {
 \includegraphics[width=\linewidth]{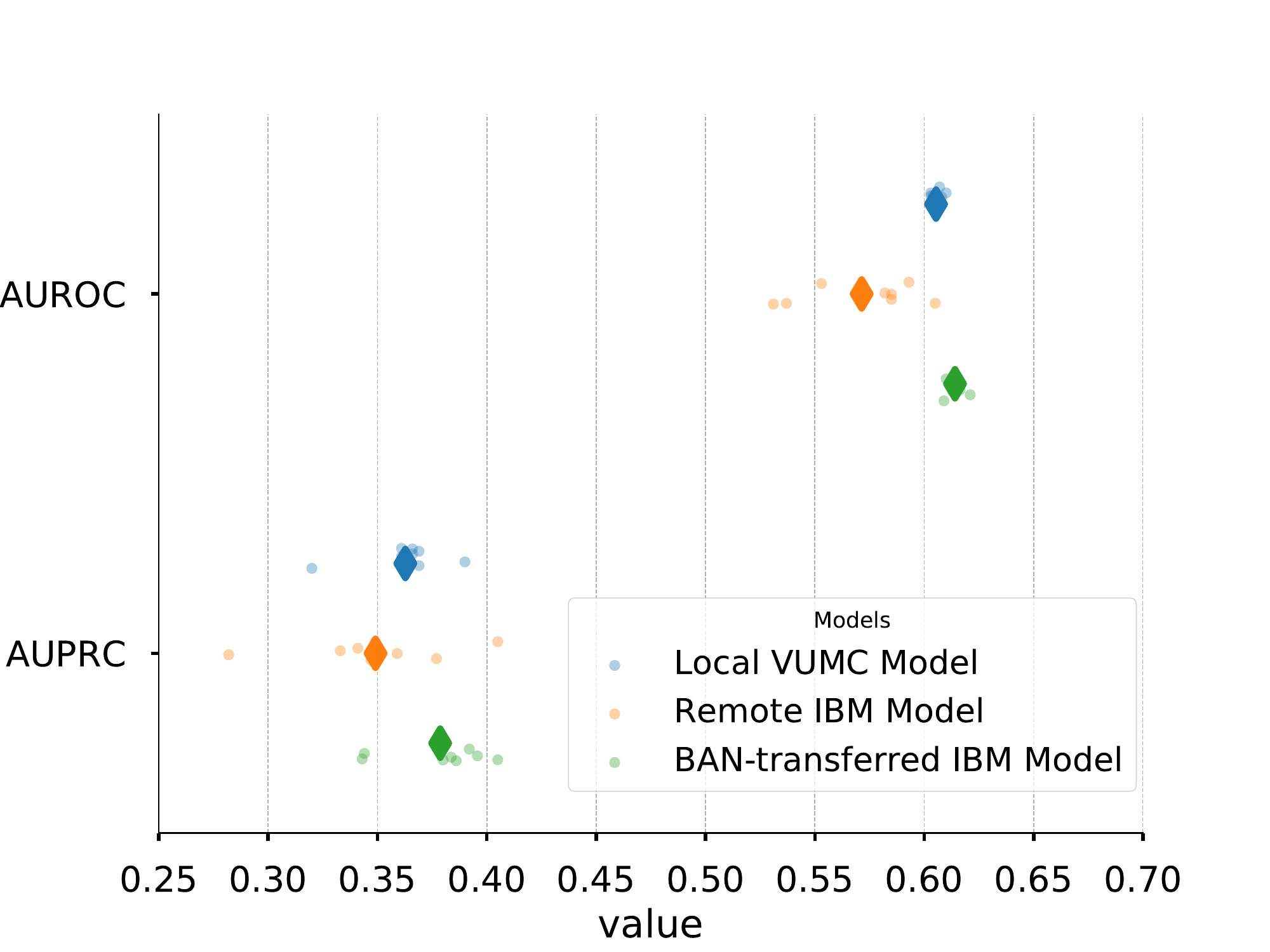}
     \caption{Performance}%
     \label{fig:direct_transfer}
    }
   \end{subfigure}
  \begin{subfigure}{.47\textwidth}
    {
     \includegraphics[width=\linewidth]{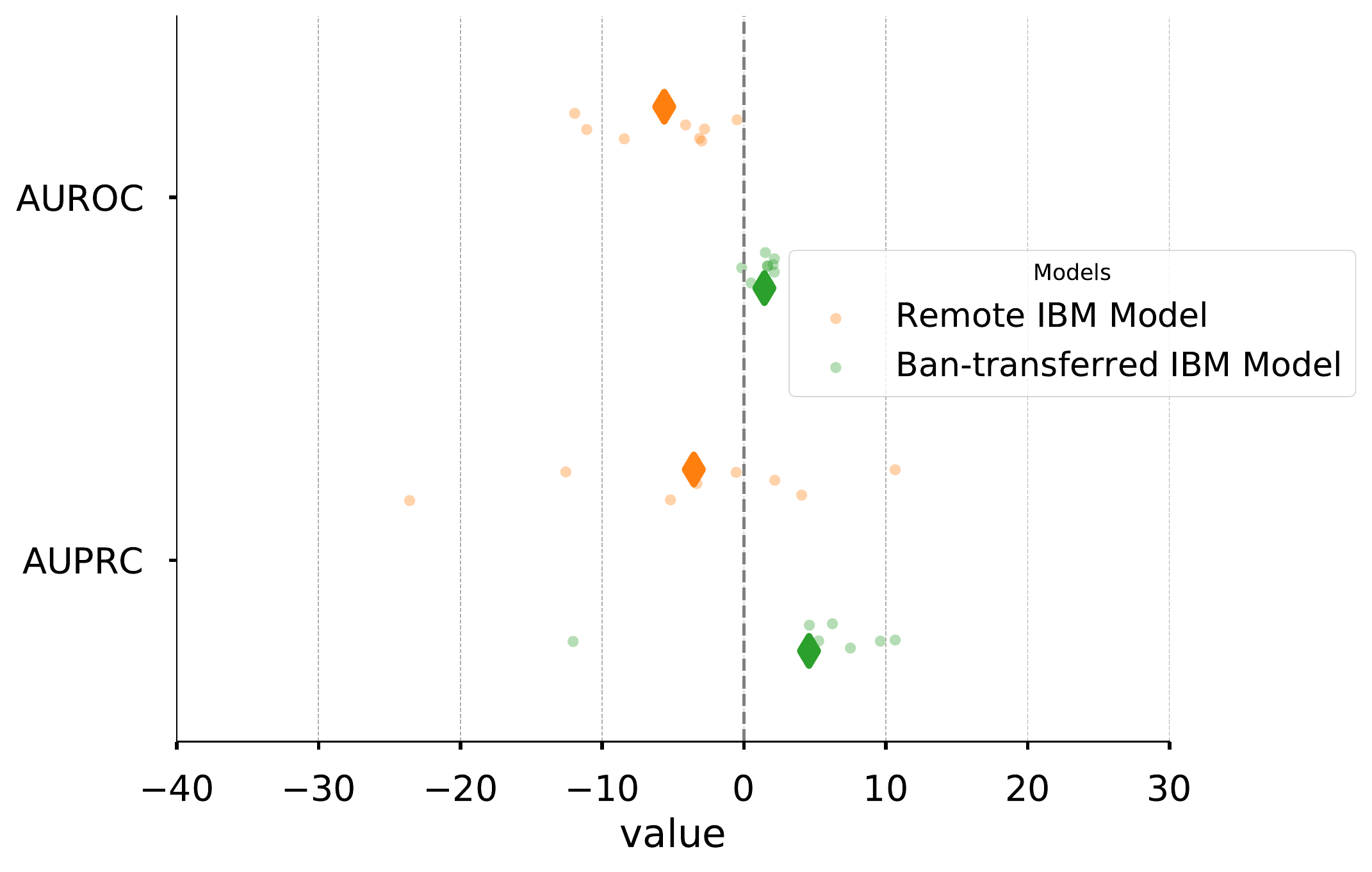}
     \caption{Lift over baseline}%
     \label{fig:consistency}
    }
  \end{subfigure}

  \caption{Performance (AUROC and AUPRC) evaluation on VUMC dataset. Comparisons shown for (a) locally built model (baseline)  against remote model and transferred remote model and (b) achieved lift in performance over baseline. Dots and diamonds represent the values for each fold and mean over all folds, respectively. 
  \label{fig:performance}}
\end{figure}

The performance of the models on each site is reported in Table~\ref{tab:perf} with respect to both area under the receiver operator characteristic curve (AUROC) and precision-recall curve (AUPRC). We report the 
performance for the best models as selected by hyperparameter selection over the validation set on the held out test set. 
Performance is reported for the local models (when applied on the same dataset as the one it was trained on), remote models (when applying it on the alternate dataset), as well as the BAN-transferred models. 
With respect to each dataset, we consider the corresponding local model as the baseline and report the change in performance as a percentage lift over the baseline. It can be seen that the directly transferred remote models generally has worse performance than the local models. 
Notably, the degradation in model performance is more pronounced for the AUPRC than the AUROC. Specifically, it can be seen that model performance is over $10\%$ worse for the directly transferred remote models with respect to AUPRC.

By contrast, the BAN-transferred models lead to slightly better AUROC performance than their respective baselines. However, more importantly, it can be seen that the AUPRC has now 
a slightly better performance 
than the baselines. This suggests that the BAN-transferred model accounts for the imbalanced data 
better than the local models.

To further investigate this improvement, we created $8$ random bootstraps of the training, validation, calibration, and test folds on the VUMC dataset 
; trained local VUMC models; and applied both the local VUMC and the remote 
IBM model on the VUMC dataset.  Figure~\ref{fig:direct_transfer} depicts the performance across these $8$ runs. 
It can be seen that, while the best runs for the remote model can achieve performance on par with the local models, on average the local models perform considerably better with respect to both AUROC and AUPRC. Interestingly, the BAN-transferred remote model leads to an improvement on average for both metrics over the local model.


Our experimental setup of $8$ distinct bootstrap folds also enables us to compare the lift achieved by BAN-transferred model over the corresponding local model (that was fine-tuned to realize the BAN-transferred model) for each run. Figure~\ref{fig:consistency} shows the performance lift and compares it against the remote model as a reference. It can be seen that the BAN-transfer process consistently increases the performance of local models (average greater than $0\%$) and this lift was found to be statistically significant for AUROC (0.05 significance level for paired t-test with two side alternate hypothesis). Furthermore, $87.5\%$ of the bootstrap runs led to a lift for AUPRC. These show the consistency of the transfer procedure in improving a locally built model. 

\section{Discussion and Conclusions}

This paper presents a new method of transferring models across sites using a computationally cheap process by only considering the models as black-box. 
The empirical findings
suggest that a BAN-based transfer process shows promise in transferring ML models across healthcare organizations.  This approach is notable in that allows the locally learned models to be treated as a black-box. 
%
In future work, we plan to investigate various experimental settings to ground the improvements in BAN-transfer and define the boundaries when such transfer is successful.

\section{Acknowledgments}

This work was funded in part by IBM Watson Health and NIH grant ULTR002243.

\bibliographystyle{ACM-Reference-Format}
\bibliography{sample-base}

\end{document}